\documentclass[conference]{IEEEtran}
\IEEEoverridecommandlockouts

\usepackage[backend=bibtex,style=numeric, maxnames=5, sorting=none]{biblatex}
\usepackage{amsmath,amssymb,amsfonts}
\usepackage{algorithmic}
\usepackage{graphicx}
\usepackage{textcomp}
\usepackage{xcolor}

\usepackage{subcaption}
\usepackage{tabularx}
\newcolumntype{P}[1]{>{\centering\arraybackslash}p{#1}}

\usepackage{makecell}
\usepackage[rightcaption]{sidecap}

\usepackage[font=small,skip=5pt]{caption}

\usepackage{pdfpages}

\def\BibTeX{{\rm B\kern-.05em{\sc i\kern-.025em b}\kern-.08em
    T\kern-.1667em\lower.7ex\hbox{E}\kern-.125emX}}

\newif\ifanonymized
\anonymizedfalse   

\newcommand{\anon}[1]{\ifanonymized\else#1\fi}

\begin{document}
	
\includepdf{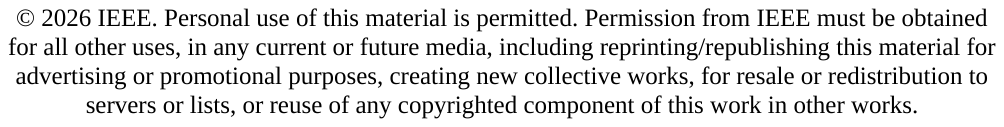}

\title{Deep Neural Network Based Roadwork Detection for Autonomous Driving}

\anon{\author{
	Sebastian Wullrich\textsuperscript{1}, Nicolai Steinke\textsuperscript{1} and Daniel Goehring\textsuperscript{1}
}}

\maketitle

\begingroup
\renewcommand\thefootnote{\textsuperscript{1}}
\anon{\footnotetext[1]{{\scriptsize\parbox{\linewidth}{Dahlem Center for Machine Learning and Robotics, Freie Universität Berlin}}}}

\endgroup

\begin{abstract}
Road construction sites create major challenges for both autonomous vehicles and human drivers due to their highly dynamic and heterogeneous nature. This paper presents a real‑time system that detects and localizes roadworks by combining a YOLO neural network with LiDAR data. The system identifies individual roadwork objects while driving, merges them into coherent construction sites and records their outlines in world coordinates. The model training was based on an adapted US dataset and a new dataset collected from test drives with a prototype vehicle\anon{ in Berlin, Germany}. Evaluations on real-world road construction sites showed a localization accuracy below 0.5 m. The system can support traffic authorities with up-to-date roadwork data and could enable autonomous vehicles to navigate construction sites more safely in the future.
\end{abstract}
\section{Introduction}
Road construction sites pose significant challenges for both human drivers and autonomous vehicles due to their dynamic layouts, heterogeneous appearance and frequently missing or unclear lane markings. This increases cognitive load for drivers and complicates perception for autonomous systems. Although roadworks account for a small fraction of roadway usage, US data show that they cause around 24\% of non‑recurring delays and 482 million vehicle hours lost annually \cite{delays_US} and are associated with numerous accidents: in 2022 alone, 891 fatalities and 37,701 injuries occurred in US roadwork zones \cite{work_zone_crashs}. Current driver-assistive and autonomous systems often fail in these complex environments, highlighting the need for reliable, real-time detection and localization of roadwork elements, ideally at around 30 frames per second (fps) on moderate hardware. This work presents a framework that fuses camera and LiDAR data to recognize and track roadwork objects. To achieve this, an existing US dataset was adapted and relabeled for German traffic and used for pre-training the final YOLO model, while a newly created dataset of German roadwork objects was used for fine-tuning. The system is optimized for the prototype vehicle ``MadeInGermany'' \cite{prototype_vehicle} by AutoNOMOS Labs, Freie Universität Berlin. This paper presents a practical framework for real-time roadwork detection using camera and LiDAR inputs, describes the adaptation of a US roadwork dataset for German traffic and the creation of a new German roadwork dataset, and analyzes the system’s performance and limitations. The developed system could also help urban authorities monitor road construction sites continuously and update maps in a timely manner.
\section{Related Work}
Roadworks are diverse and less common than standard traffic scenarios, making accurate annotation challenging. They are underrepresented in major datasets like BDD100k \cite{BDD100K} and Mapillary \cite{mapillary}, where only common elements like cones are labeled. The ROADWork dataset \cite{american_roadwork_dataset} addresses this gap, but rare configurations remain scarce, suggesting a need for supplementation with synthetic or web data. ROADWork uses Mask R-CNN \cite{Mask-RCNN} for pixel-level localization, but at only around 5 fps on a Nvidia Tesla M40 GPU it is unsuitable for real-time use. Sundharam et al. \cite{Segmentation_of_Work-Zone_Scenes} apply ResNet and U-Net for pixel-level segmentation, potentially in real-time, aiming for Advanced Driver Assistance Systems (ADAS) integration. Abodo et al. \cite{SHRP_2_NDS} developed a deep CNN classifier via active learning to efficiently label roadwork scenes, while Mathibela et al. \cite{anticipate_roadworks} fused uncertain priors (from web sources) with onboard observations (e.g. cones, signs) using HMM and Gaussian Processes to estimate the likelihood of roadworks along a route. Kunz and Schreier \cite{Construction_Sites_on_Motorways} used a Bayesian network that fuses uncertain cues from onboard sensors to detect motorway construction sites without map priors, reliably alerting drivers. Seo et al. \cite{Highway_Workzones} used a forward-facing camera with an AdaBoost orange-pixel detector and SVM to identify roadwork signs, inferring boundaries indirectly from start/end sign detections rather than measuring the blocked area directly. Most of these approaches rely solely on cameras, limiting accurate boundary localization. To overcome this limitation, Shi and Rajkumar \cite{work_zone_detection} proposed a three-stage pipeline that detects keypoints on roadwork objects, projects them into a bird’s-eye view and clusters them with convex hulls using DBSCAN, a density-based clustering algorithm, combining EfficientDet-D0 image detection with 3D LiDAR keypoints. Their method is limited by the flat-road assumption, which can distort boundaries on slopes or uneven terrain; by a reported maximum detection range of 27 m, making real-time safety at speeds above 50 km/h uncertain; and by an unknown runtime. In contrast, this paper uses a faster YOLO network (YOLO11m \cite{yolo11_ultralytics}) with low-resolution Ibeo LUX LiDAR sensors. The simpler and more cost-effective hardware setup enables faster and longer-range localization of roadwork objects (at least 50 m) and offers greater robustness on sloped or uneven road surfaces.
 
\section{Implementation}
\subsection{Datasets}
\subsubsection{ROADWork Dataset (US)} \label{roadwork_dataset}
\begin{figure}[tb]
	\centering
	\begin{subfigure}[b]{0.24\textwidth}
		\centering
		\includegraphics[width=0.72\textwidth]{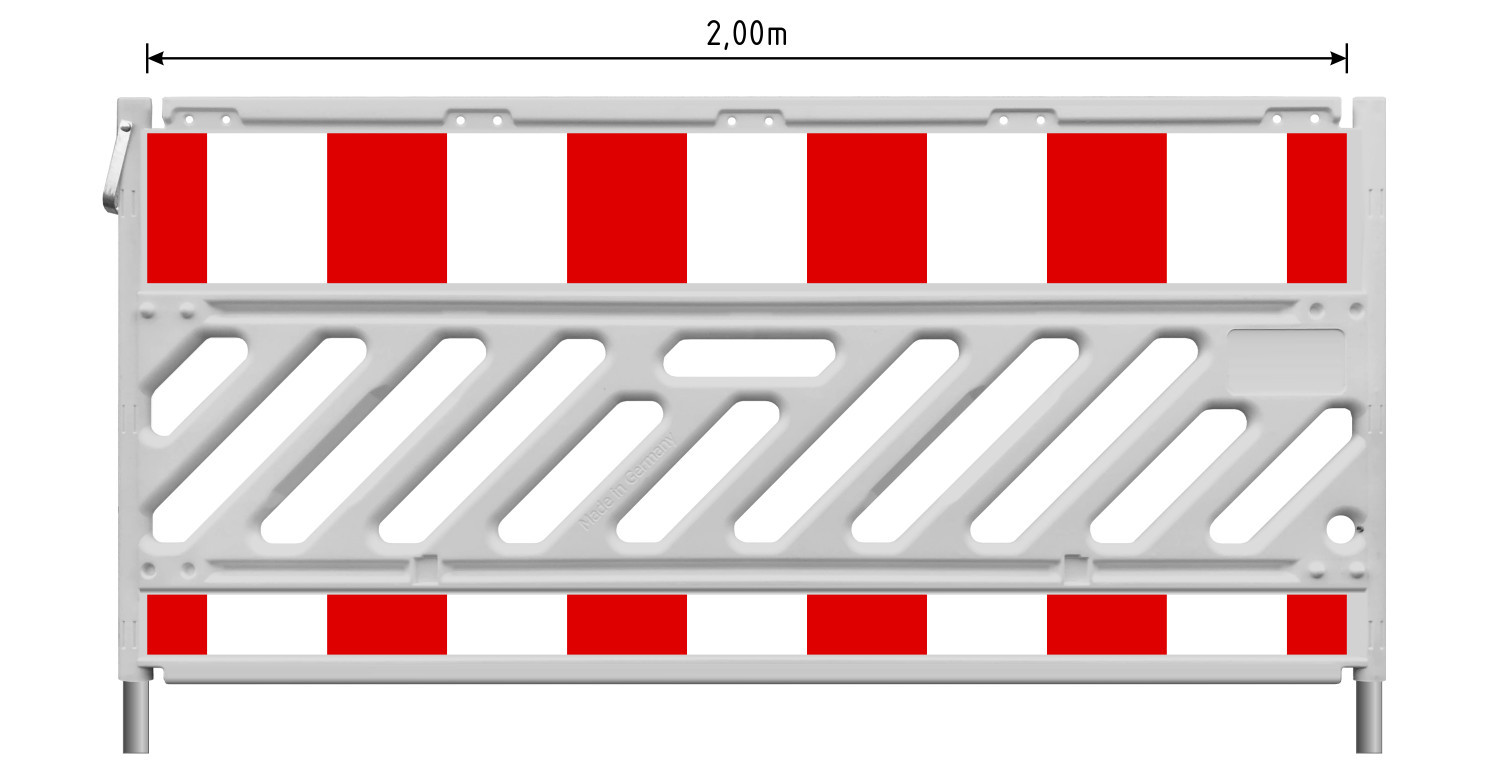}
		\caption{}
		\label{barrier}
	\end{subfigure}
	\hfill
	\begin{subfigure}[b]{0.24\textwidth}
		\centering
		\includegraphics[width=0.17\textwidth]{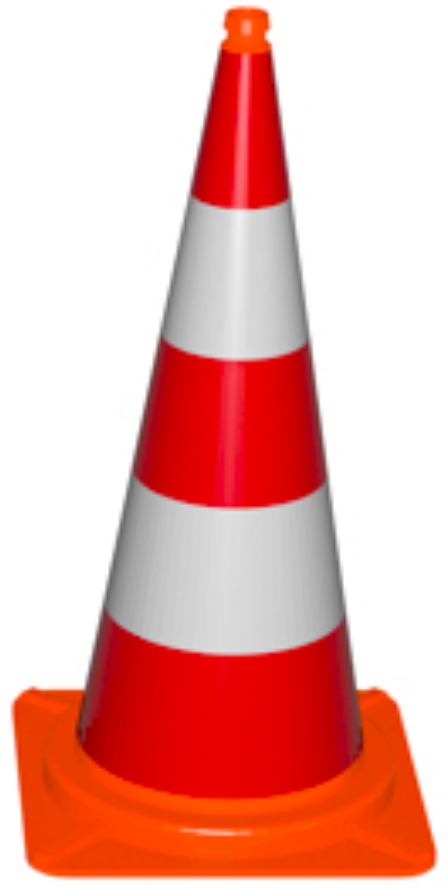}
		\caption{}
		\label{traffic_cone}
	\end{subfigure}
	
	\begin{subfigure}[b]{0.24\textwidth}
		\centering
		\includegraphics[width=0.43\textwidth]{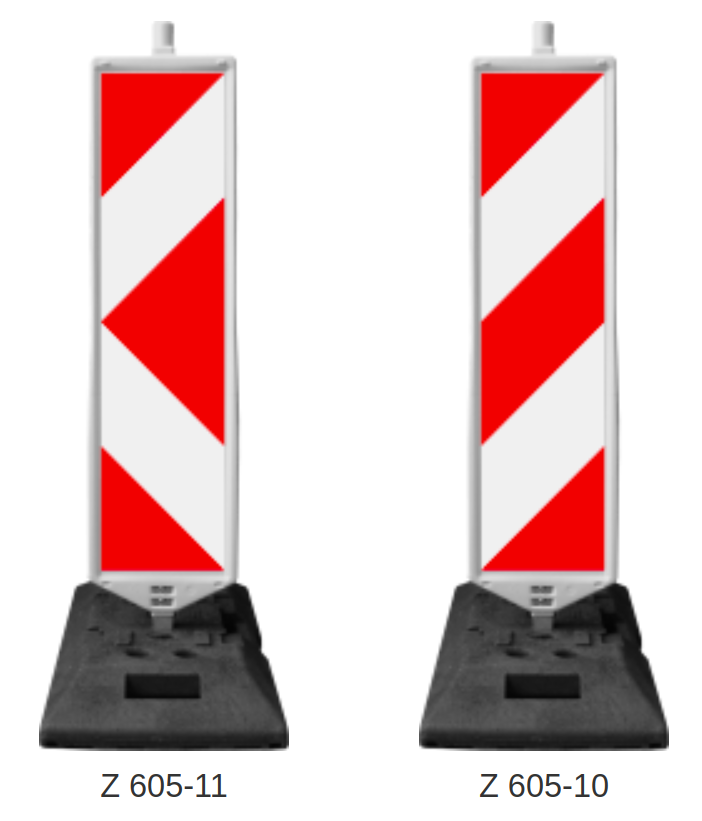}
		\caption{}
		\label{pass_left}
	\end{subfigure}
	\hfill
	\begin{subfigure}[b]{0.24\textwidth}
		\centering
		\includegraphics[width=0.41\textwidth]{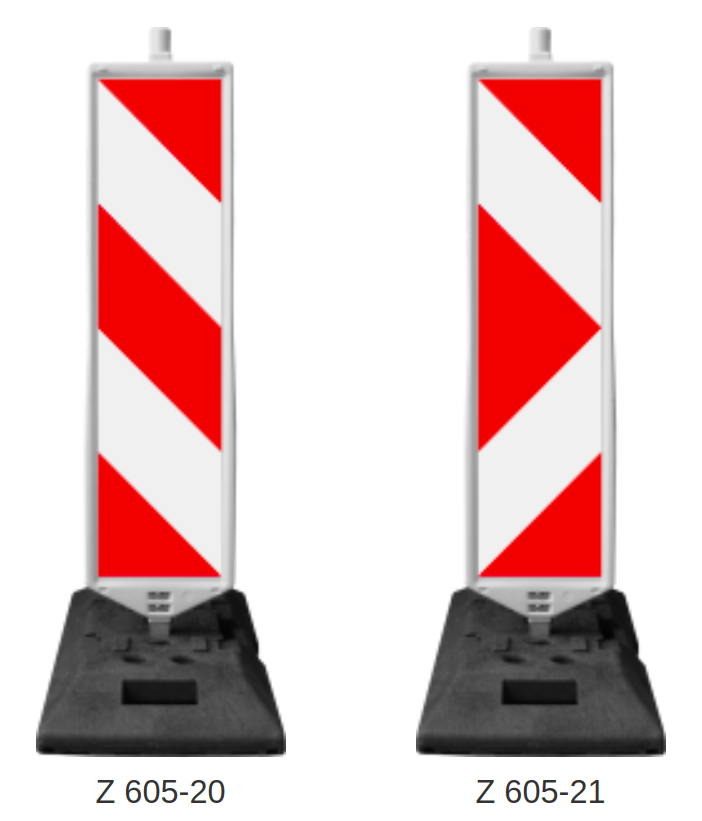}
		\caption{}
		\label{pass_right}
	\end{subfigure}
	\caption{Roadwork objects indicating road construction sites in Germany
	\cite{rsa_mainpage}: (a) Barrier; (b) Traffic cone; (c) Left-facing vertical panel (``pass left''); (d) Right-facing vertical panel (``pass right'')}
	\label{german_roadwork_classes}
\end{figure}

The ROADWork dataset \cite{american_roadwork_dataset} contains over 7,000 high-resolution images with more than 5,000 annotated roadworks, collected in various US cities. It covers 15 object categories commonly found at construction sites, but most are not suitable for this work, as they either do not define the boundaries of roadworks or do not conform with German roadwork standards. After cleaning and re-annotation, a new dataset was created, focusing on only classes relevant to Germany (see Fig. \ref{german_roadwork_classes}), totaling \url{~}27,500 instances in \url{~}4,200 images. Traffic cones disproportionately dominate the dataset (68\%) because many frames were pre-selected by a cone detector.
\subsubsection{\anon{AutoNOMOS Labs} Dataset} \label{AutoNOMOS_Labs_dataset}
A second dataset was generated from test drives of the prototype autonomous vehicle ``MadeInGermany'' \cite{prototype_vehicle}. Through manual filtering, only the frames depicting roadworks (preferably from multiple angles) were retained. The final set includes ~2,000 images with over 10,500 annotated instances. Traffic cones are rarely used in German roadworks, as they are typically reserved for temporary incidents. This is reflected by the mere 89 instances of this class in the \anon{AutoNOMOS} dataset. Additional images of visually similar non-roadwork objects were added to reduce false positives.
\subsection{YOLO Training Setup and Hyperparameter Settings}
YOLO11m by Ultralytics \cite{yolo11_ultralytics}, pre-trained on the COCO dataset, was chosen for its balance of accuracy, speed and model size, avoiding overfitting risk on the relatively small ROADWork and \anon{AutoNOMOS Labs} datasets, compared to larger model variants. Models were trained for up to 600 epochs with early stopping after 100 stagnant epochs, using an image size of 640 $\times$ 640, a batch size of 8 and a cosine learning rate schedule starting at $10^{-4}$ and decaying to $10^{-6}$. Extensive data augmentation (hue, saturation, brightness shifts, rotations, shear, scaling, translation, mosaic augmentation and random erasing) during training helped the model handle diverse conditions and occlusions and the AdamW optimizer improved generalization.
\subsection{Technology Infrastructure} \label{sec:tech_infrastructure}
Data was collected using the \anon{``MadeInGermany'' research vehicle \cite{prototype_vehicle}, a modified VW Passat} equipped with vision and LiDAR sensors. A custom RCCB camera with an Onsemi AR0220 sensor captured images at 1820 $\times$ 940 pixels, averaging around 20 fps. Six Ibeo LUX LiDAR sensors provided object contours, each with 8 laser channels (low-resolution), a sensing range of up to 200 m and achieving \url{~}10 cm accuracy. For ground truth, a Velodyne Alpha Prime 128-beam 3D LiDAR offered dense 360° point clouds at 0.2° $\times$ 0.1° resolution, detecting targets up to 245 m. In the relevant 15 m range of the evaluation, this yields a spatial resolution of roughly 5.2 $\times$ 2.6 cm. The system software was built with ROS 2 and implemented in Python. Model training, evaluation and the roadwork detection experiments were carried out on a laptop with an Intel Core i9-9900K CPU and an NVIDIA GeForce RTX 2080 GPU (CUDA 12.8) running Ubuntu 24.04.
\subsection{System Implementation}
\begin{figure}[tb]
	\centering
	\includegraphics[width=0.49\textwidth]{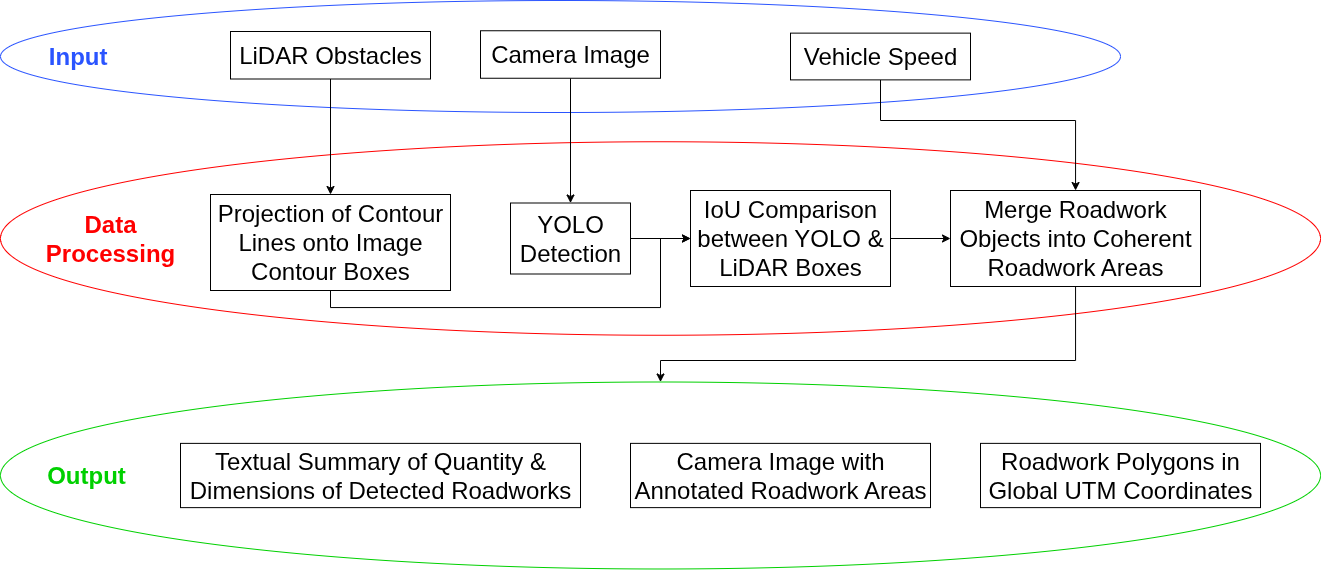}
	\caption{Dataflow of the roadwork detection system}
	\label{system_architecture}
\end{figure}

The system detects and reports roadworks in real-time using data from multiple sensors. It has three outputs: (1) camera images annotated with roadwork areas obtained by combining multiple CNN-predicted bounding boxes of roadwork objects, (2) a textual summary indicating whether roadworks are present and, if so, their number and sizes, and (3) recorded roadwork polygons in global UTM coordinates. The system receives data from three sources: (1) image data from the camera, which is used for the predictions by the YOLO CNN, (2) obstacle contour lines from the LUX LiDAR sensors for localization and tracking and (3) vehicle odometry data to calculate speed. For a better overview, the system dataflow is illustrated in Fig. \ref{system_architecture}.
\subsubsection{Odometry Processing}
The vehicle’s current speed is used to dynamically adjust the detection threshold (the number of times a roadwork object must be identified before it is considered as part of a detected roadwork). At higher speeds, fewer frames are available, so the threshold is lowered accordingly. This is done using the formula:
\begin{align*}
	\text{T}(v) &= \text{round}\left(a \cdot \ln\left(\frac{\frac{d}{v} \cdot \text{FPS}}{b}\right)\right) \quad | a = 5 \land b = 12.5
\end{align*}
\begin{align}
	\text{Detection\_Threshold} &= 
	\left\{
	\begin{array}{ll}
		2 &, \text{if} \; \text{T}(v) < 2 \\
		\text{T}(v) &, \text{if} \; 2 \leq \text{T}(v) \leq 5 \\
		5 & ,\text{if } \; \text{T}(v) > 5
	\end{array} \label{detection_threshold}
	\right.
\end{align}
Here, \textit{d} (scanner range) is limited to 50 m (see Section \ref{implementation_hella}) and an expected processing rate of FPS = 10 is used (see Section \ref{system_performance}). The parameters \textit{a} = 5 and \textit{b} = 12.5 were chosen to yield thresholds of 5 at 50 km/h, 3 at 80 km/h and 2 at 100 km/h. A logarithmic function ensures the threshold decreases faster at higher speeds and it is limited between 2 and 5 to balance detection reliability and avoid false positives.
\subsubsection{LiDAR Processing}
Detected obstacle contour points are first transformed from the robot coordinate system to the camera system and then projected into image pixel coordinates. These contour points are not raw LiDAR measurements but are generated by the LUX system, which consists of multiple 8-line LiDARs connected to an Electronic Control Unit (ECU). The ECU fuses the point clouds from all sensors and performs built-in object detection and tracking, producing structured object data including ordered contour points. The measurements from all six sensors are additionally transformed into a common robot coordinate system. Due to the low and flat scanning pattern of the LUX LiDARs, obstacle height cannot be directly measured. Therefore an upper contour is estimated assuming a standard object height of 1.60 m, covering the tallest objects (vertical panels up to 1.3 m plus an optional 0.3 m warning light). These top and bottom lines are combined to create the full contour box. A special case occurs with long objects (e.g. barriers) that extend beyond the camera’s field of view. In these situations, a new point is interpolated at the image boundary so that the object remains fully represented within the image. Finally, contour points are transformed into a local world coordinate system. Its origin is fixed at the vehicle’s position at the start of a drive and remains constant for the entire session, which helps keep coordinate values small and minimizes rounding errors compared to using the UTM system.
\subsubsection{Camera Processing} \label{implementation_hella}
\begin{SCfigure}
	\centering
	\includegraphics[width=0.2\textwidth]{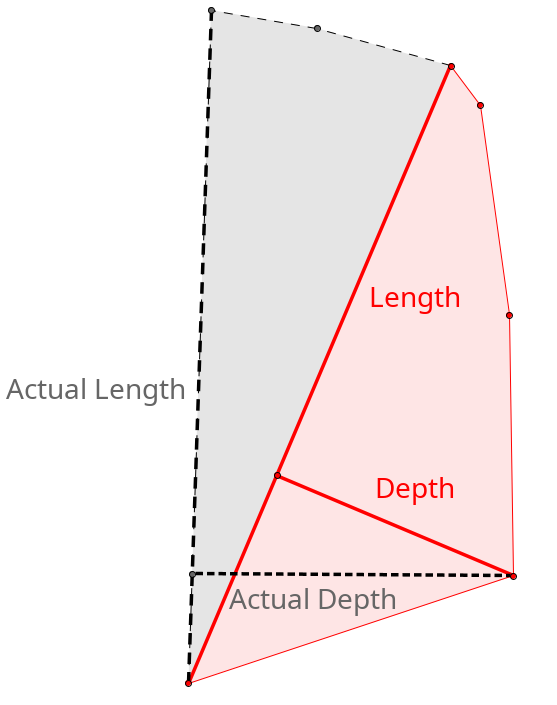}
	\caption{Red area: illustrates the current measurement of roadwork dimensions utilizing detected roadwork objects as the vehicle passes (moving from bottom to top, passing on the right); Gray area: undetected portion representing the full extent of the actual roadwork if all its objects were detected}
	\label{Tiefenberechnung}
\end{SCfigure}

\begin{figure}[tb]
	\centering
	
	\begin{subfigure}[b]{0.45\textwidth}
		\centering
		\includegraphics[width=0.8\textwidth]{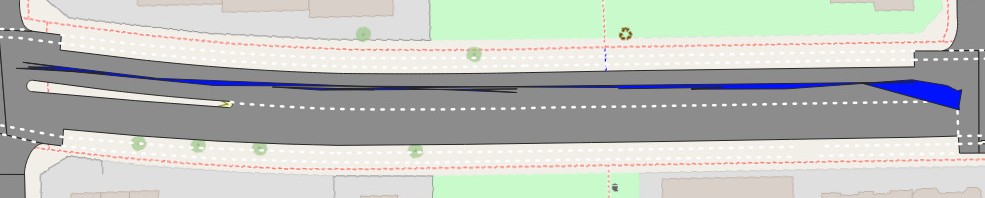}
		\caption{}
		\label{roadwork_smootherer_2_a}
	\end{subfigure}
	
	\begin{subfigure}[b]{0.45\textwidth}
		\centering
		\includegraphics[width=0.8\textwidth]{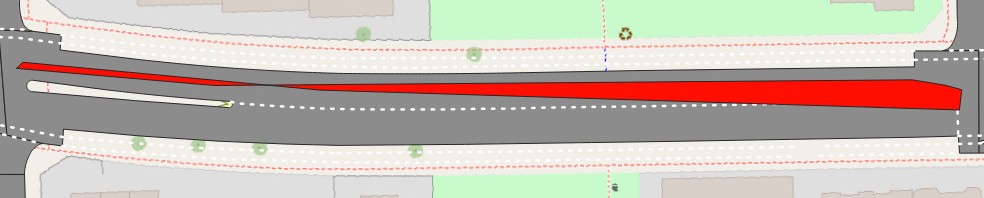}
		\caption{}
		\label{roadwork_smootherer_2_b}
	\end{subfigure}
	
	\caption{Differing shapes of final roadwork outputs: (a) Unaltered version recorded during driving; (b) Refined version using only convex hull outer points}
	\label{roadwork_smootherer}
\end{figure}

The system processes the camera frames (originally 1820 $\times$ 940 px, downscaled to 640 $\times$ 352 px) only when new LiDAR data is received, ensuring that current object contours are always processed. The YOLO11m CNN detects roadwork objects with high confidence thresholds (0.75 for barriers, 0.7 for vertical panels and cones, see Section \ref{eval_cross_validation}). If no roadwork is detected, the resized images are output unchanged. To identify actual roadwork objects, predicted CNN bounding boxes are matched to LiDAR-derived contour boxes. A match is indicated if the Intersection over Union (IoU) exceeds 50\%. When multiple candidates exceed this threshold, instead of simply choosing the contour with the highest IoU (which tends to favor larger objects) the candidate is chosen whose LiDAR object line in image pixels is closest to the bottom edge of the CNN bounding box. This alignment ensures more accurate object matching. Objects whose LiDAR contour boxes are over twice as large as the CNN prediction (for vertical panels or cones) are excluded to avoid them merging with walls. Barriers are exempt as they are often connected and therefore form long continuous contours. An object tracker keeps a dictionary of currently detected roadwork objects (class, contour, detection count). To avoid false associations with distant walls that appear narrow in the contour data and therefore bypass size filtering, tracking is restricted to objects within 50 m. If the CNN temporarily misses objects (e.g. due to occlusion in specific frames), missing LiDAR-detected objects are still added, provided they meet the detection threshold (Eq. \ref{detection_threshold}). The tracked roadworks are stored in a hierarchical dictionary, with the individual tracked roadworks at the top level, while the lower levels contain associated obstacle IDs of the roadwork objects, along with their corresponding contour points in local world coordinates. The first detected object stores its full contour (needed to measure the roadworks depth), while subsequent objects retain only their last contour point to keep the site outline smooth. Objects are assigned to the same roadwork if their mutual separations do not exceed the following thresholds:\vspace{0.2cm}\\
\textbf{Longitudinal separation:}
\begin{itemize}
	\item \textbf{Vertical-panel-to-Vertical-panel:}
	12 m (based on a typical 10 m spacing in Germany plus a safety buffer) \cite{rsa_allgemeines}.
	\item \textbf{Barrier‐to‐barrier:}
	2 m (to ensure continuous enclosure of the work zone).
	\item \textbf{Barrier-to-other‐object:}
	6 m (in accordance with urban spacing guidelines in Germany dictating 5 m) \cite{rsa_allgemeines}.
\end{itemize}
\textbf{Lateral separation:} 1.5 m (slightly exceeding the 1 m maximum lateral spacing between elements in Germany) \cite{rsa_mainpage}.
\vspace{0.2cm}\\
If a newly detected object meets the above measurements relative to an object from an existing roadwork, it is inserted into the correct position within this site. If it becomes the new first object of this roadwork, its full contour is stored, while the contour of the previous first object is trimmed to its last point. Sometimes, longer roadworks may temporarily split into two segments if intermediate objects have not yet been detected often enough. This issue is identified when a roadwork object appears as part of two different construction sites in the roadwork dictionary. It is resolved by merging all objects from both sites into a single roadwork, based on their relative distances. Detected sites are visualized by drawing them in the camera image. ``Ghost objects'' are added when roadwork objects move out of view but are expected to remain relevant as part of the same site over the next few meters, thereby maintaining the roadwork’s visual continuity. These do not introduce new measurements but serve only as a visual aid to support the correct recognition and continuous representation of the road construction site. Finally, smaller roadworks detected within larger ones are removed by retaining only the largest. The roadwork length is estimated as the maximum Euclidean distance between the first point and all other points, and the depth is defined as the maximum orthogonal distance from this axis (see Fig. \ref{Tiefenberechnung}).
If no roadwork objects are detected for 50 m after the last object, the roadwork is considered finished. Results are converted and saved in global UTM coordinates and the object tracker and roadwork dictionary reset. For output, both a raw polygon (captured during driving) and a smoothed version (using a convex hull algorithm) are saved. The latter better approximates real-world shapes of road construction sites while sacrificing detail (see Fig. \ref{roadwork_smootherer}).
\section{Experiments / Evaluation}
\subsection{YOLO Model}
\subsubsection{Model Improvement via Background Images}
One key observation during training was that using only 10\% background images per dataset led to a relatively high number of false positives. To address this, a model initially trained with 10\% background images was evaluated on over 100 rosbag files, corresponding to approximately 20 hours of recorded test drives. In the roughly 100,000 frames where the model produced predictions, each prediction was manually reviewed and categorized based on the actual presence or absence of roadwork objects. This analysis identified more than 3,200 problematic frames, which were then proportionally added as background images to the training, validation and test sets. As a result, the final background image proportion exceeds 60\% in all three sets, meaning that there are more than 1.5 times as many background images as there are labeled images. The effectiveness of this procedure in reducing false positives was evaluated by comparing a model trained on only 10\% background images with one trained on 60\%. Both models were tested on a final set containing 60\% background images. Increasing the number of background images reduced false positives by over 80\% and additionally encouraged the network to focus on actual roadwork objects, thereby decreasing the number of false negatives by 25\%. It is important to note that the background images in the test set were never used during training, so the observed improvement in robustness is genuine.
\subsubsection{6-Fold Cross-Validation} \label{eval_cross_validation}
\begin{figure}[tb]
	\centering
	\begin{subfigure}[b]{0.24\textwidth}
		\centering
		\includegraphics[width=\textwidth]{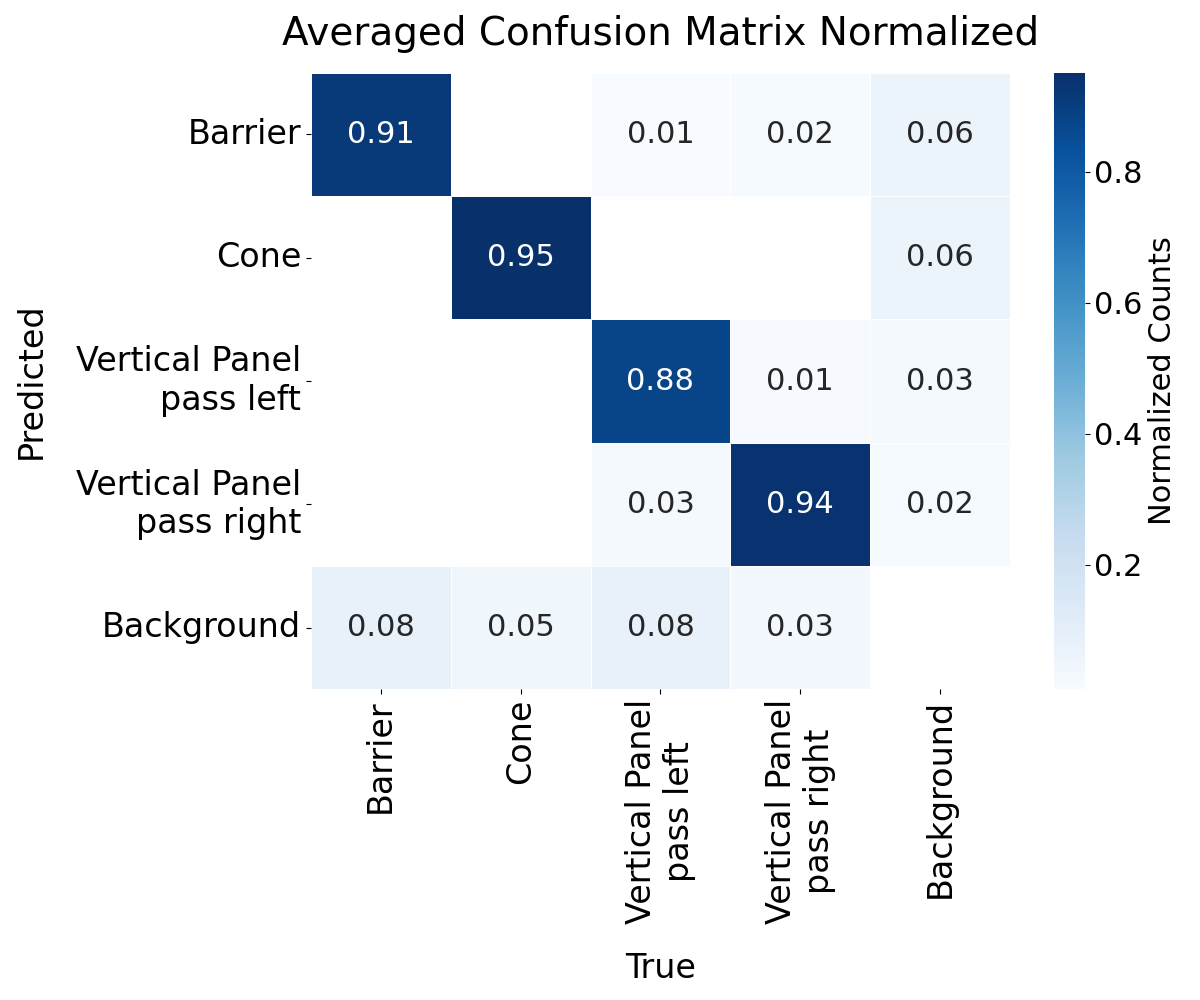}
		\caption{}
		\label{final_model_averaged_confusion_matrix_iou_0_3}
	\end{subfigure}
	\hfill
	\begin{subfigure}[b]{0.24\textwidth}
		\includegraphics[width=\textwidth]{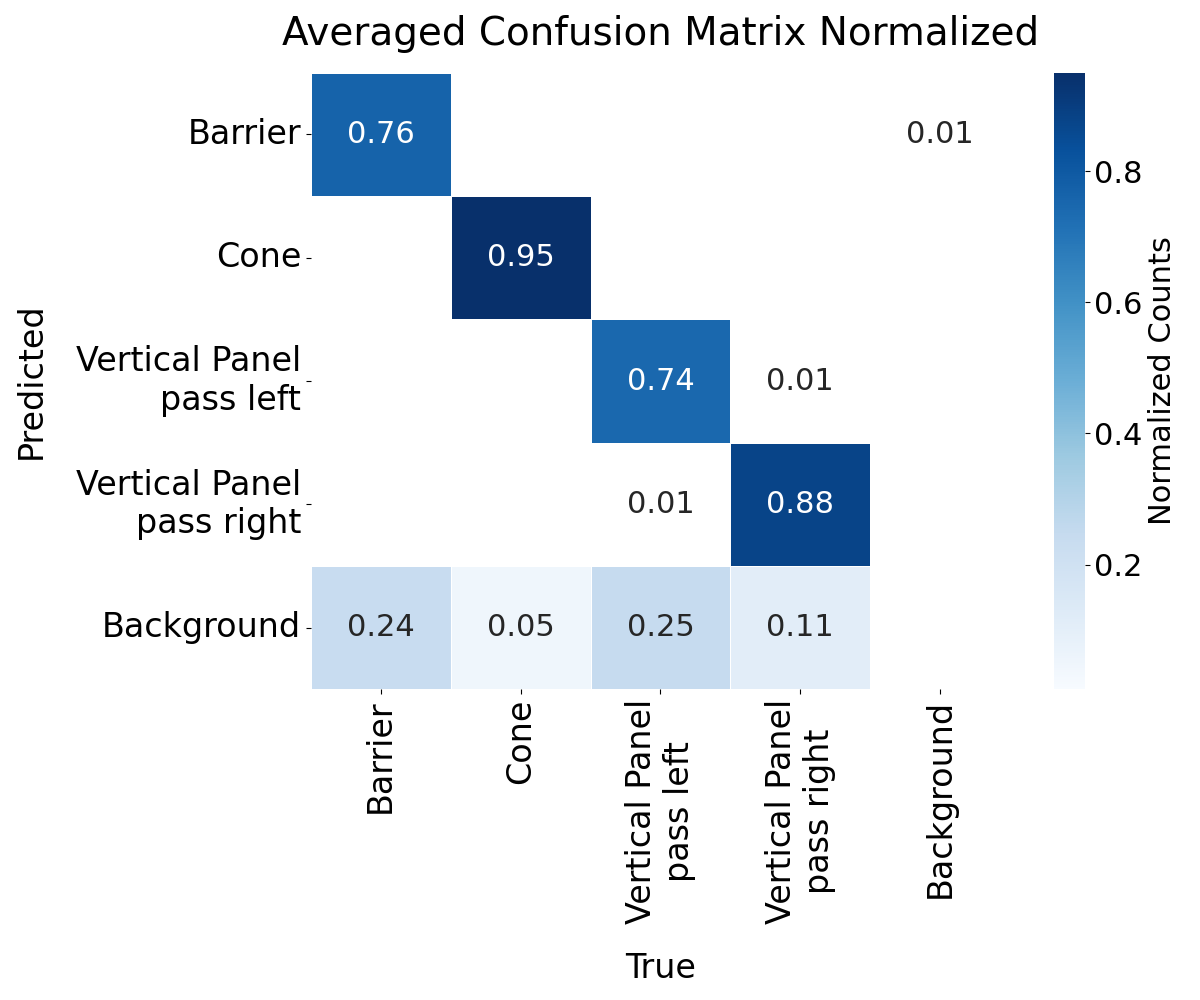}
		\caption{}
		\label{final_model_averaged_confusion_matrix_iou_0_3_my_threshold}
	\end{subfigure}
	\caption{Average confusion matrices from 6-fold cross-validation of the final YOLO model: (a) A confidence threshold of 0.001; (b) A class-specific threshold of 0.75 for barriers and 0.7 for other classes}
\end{figure}

\begin{figure}[tb]
	\centering	
	\includegraphics[width=0.32\textwidth]{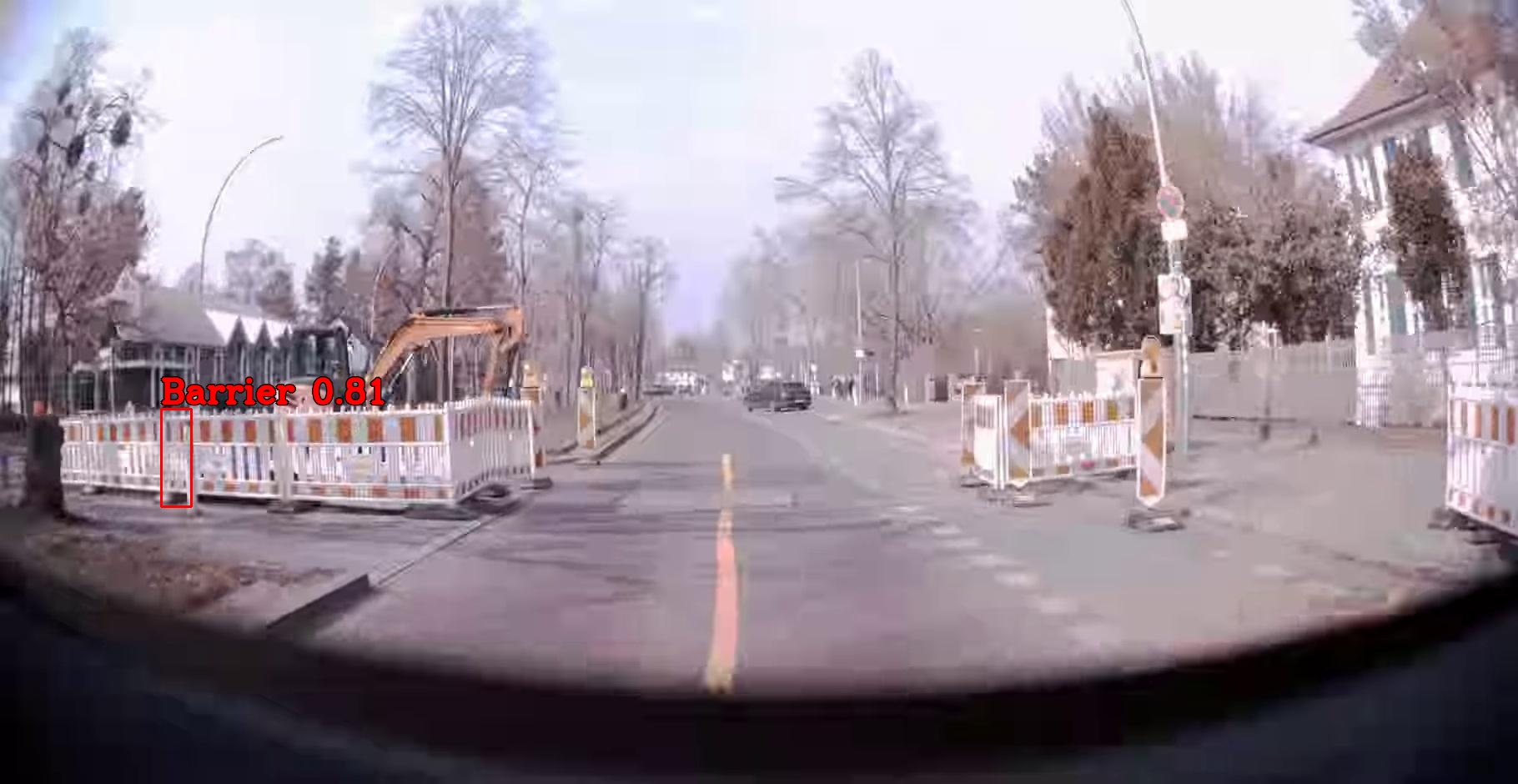}
	\caption{False positive example detected by the final model}
	\label{false_positives_final_model}
\end{figure}

\begin{figure}[tb]
	\centering
	\includegraphics[width=0.32\textwidth]{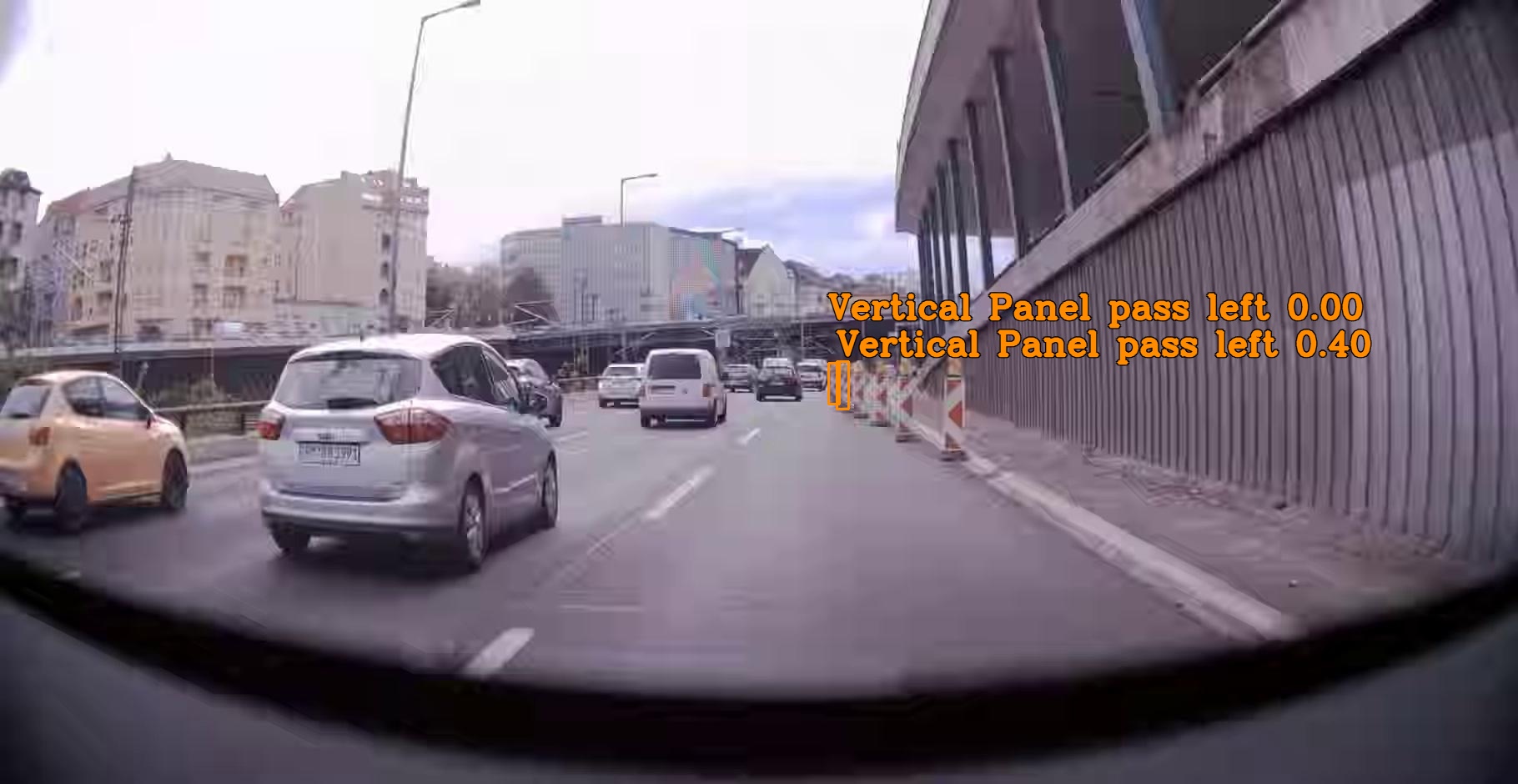}	
	\caption{False negative examples detected by the final model}
	\label{final_model_false_negative}
\end{figure}

The YOLO11m model was pre-trained on the modified ROADWork dataset from the US as described in Section \ref{roadwork_dataset} and fine-tuned on the \anon{AutoNOMOS Labs} dataset (Section \ref{AutoNOMOS_Labs_dataset}). A cross-validation was conducted in order to reliably estimate the model's performance on unseen data. Hastie, Tibshirani and Friedman \cite{cross_val_error} recommend a 5 to 10-fold cross-validation in order to balance bias and variance. Consequently, a 6-fold cross-validation was conducted, with 75\% of the dataset utilized as training data after reserving 10\% for the test set. To ensure reproducibility, a fixed seed was applied when creating the folds. The class distribution in each fold was checked beforehand and, if it deviated by more than 5\% from the original distribution, a new seed was selected and documented. The test set is additionally stratified to maintain class proportions and contains 558 barriers, 10 traffic cones, 379 ``pass left'' and 235 ``pass right'' vertical panels. The averaged confusion matrices show that the CNN achieves an average per-class recall of ~90\% across all object classes (see Fig. \ref{final_model_averaged_confusion_matrix_iou_0_3}). Occasional misclassifications (e.g. confusing left and right vertical panels, or misidentifying them as barriers) are rare and not critical, since the roadwork detection system treats these classes similarly. However, individual classes show up to 6\% false positives, which is problematic in autonomous driving, as false positives could trigger inappropriate braking. Research revealed that Ultralytics applies virtually no confidence threshold during evaluation (set to 0.001) and its default threshold for prediction is also relatively low at 0.25. F1-Confidence curves were analyzed to select thresholds that balance precision and recall. The final detection system uses a conservative threshold of 0.75 for barriers and 0.7 for vertical panels and traffic cones, prioritizing high precision to reduce false positives. Re-evaluating with these higher thresholds naturally reduced the number of true positives and increased false negatives (see Fig. \ref{final_model_averaged_confusion_matrix_iou_0_3_my_threshold}). The prediction rate for barriers and ``Vertical Panels pass left'' decreased from \url{~}90\% to \url{~}75\%, except for the ``Vertical Panels pass right'' class, which still maintains a high true positive rate of \url{~}88\%. Results for traffic cones remained largely unchanged. Crucially, as intended, false positives are now almost completely eliminated, with precision approaching 100\% across all classes (see Table \ref{class_metrics_my_conf_threshold}).
Analysis of all the remaining false positives shows that most occur within genuine roadwork scenes. For example, the false positive in Fig. \ref{false_positives_final_model} is a small barrier predicted inside a larger barrier. Only two genuine false positives remained, both at the image edge. These edge cases are not critical because detections must persist over multiple frames to be recorded as roadwork objects.
Most false negatives involve distant objects that would be detected once the vehicle approaches them (see Fig. \ref{final_model_false_negative}).
As mentioned previously, with the new confidence threshold precision is close to 100\%, while recall has declined to around 75\% for two of the four classes (see Table \ref{class_metrics_my_conf_threshold}). Accuracy and the F1-score are near 90\% for all classes.
\begin{table}[tb]
	\caption{Mean and standard deviation per class of the different metrics across all six folds of the cross-validation of the final YOLO model with the new confidence thresholds}
	\label{class_metrics_my_conf_threshold}
	\centering
	\resizebox{0.5\textwidth}{!}{%
		\begin{tabular}{l|c|c|c|c}
			\textbf{Metric} & \textbf{Barrier} & \textbf{Cone} & \textbf{Vertical Panel pass left} & \textbf{Vertical Panel pass right} \\
			\hline
			Accuracy   & $0.885 \pm 0.015$ & $1.000 \pm 0.000$ & $0.916 \pm 0.010$ & $0.974 \pm 0.004$ \\ 
			Precision  & $0.990 \pm 0.005$ & $1.000 \pm 0.000$ & $0.992 \pm 0.003$ & $0.989 \pm 0.005$ \\ 
			Recall     & $0.763 \pm 0.030$ & $0.950 \pm 0.112$ & $0.745 \pm 0.031$ & $0.880 \pm 0.016$ \\ 
			F1-score   & $0.861 \pm 0.020$ & $0.971 \pm 0.066$ & $0.850 \pm 0.020$ & $0.931 \pm 0.010$ \\
		\end{tabular}
	}
\end{table}
A major limitation is the small number of traffic cone instances (only 89 overall).
However, because the model was pre-trained on the ROADWork dataset, which contains over 18,500 cones, it is still capable of accurate traffic cone detection. Furthermore, traffic cones are less important in German roadworks as they are rarely used and only in temporary situations that are expected to be resolved quickly. The standard deviation (SD) across folds is low, except for recall (and consequently the F1-score) on cones, where one fold failed to detect three cones behind a mesh fence. This is a rare scenario and all other folds detected these cones correctly. The consistent performance across folds and low SD indicate that the model’s quality is not due to chance in data splits. Overall, the ROADWork and \anon{AutoNOMOS Labs} datasets, combined with Ultralytics’ YOLO11m, yielded a robust model capable of reliably detecting and localizing roadwork objects. False positives have been almost completely eliminated and recall remains relatively high ($\geq$ 75\%) even at these high confidence thresholds, with most undetected objects being too far away. Additionally, processing a multi-frame video stream further increases the likelihood of correctly identifying roadwork objects. The concern that entire roadworks could go undetected is unfounded, as demonstrated in Section \ref{evaluation_roadworks}, where the YOLO model successfully detected all relevant objects at real-world road construction sites.
\subsection{System Measurement Accuracy Evaluated on Real-World Roadworks} \label{evaluation_roadworks}
\begin{figure}[tb]
	\centering
	\begin{subfigure}[b]{0.49\linewidth}
		\centering
		\includegraphics[width=0.69\linewidth]{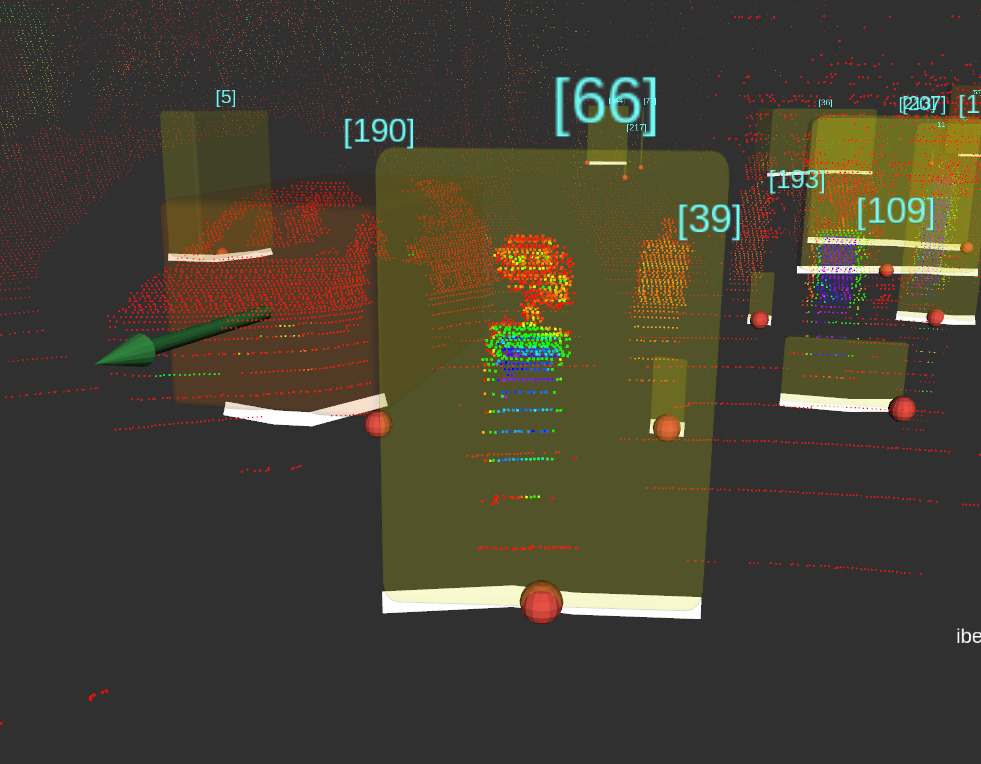}
		\caption{}
		\label{Panels_too_small_in_ibeo}
	\end{subfigure}
	\hfill
	\begin{subfigure}[b]{0.49\linewidth}
		\centering
		\includegraphics[width=0.4\linewidth]{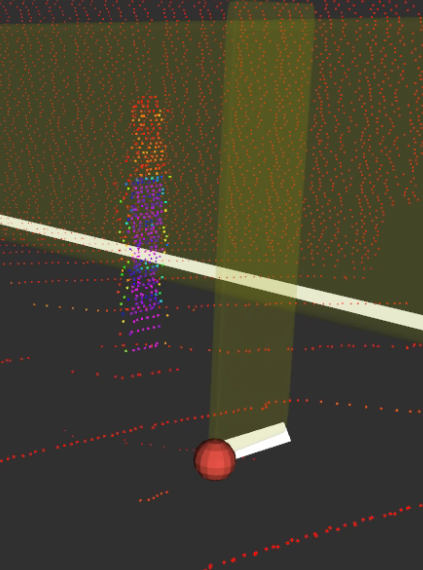}
		\caption{}
		\label{ibeo_versetzt}
	\end{subfigure}
	\caption{Examples where the LUX sensor contours differ from the Velodyne point cloud}
	\label{Standfuesse}
\end{figure}

\begin{figure}[tb]
	\centering
	\includegraphics[width=0.25\textwidth]{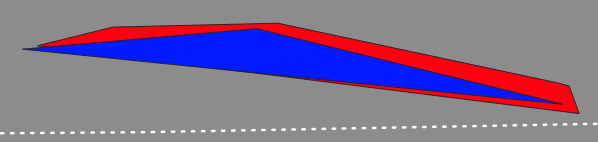}
	\caption{Comparison between system measurements (red) and ground truth (blue)}
	\label{GT_test}
\end{figure}

It should be noted that all roadworks evaluated in the following analysis were deliberately excluded from the training and validation sets, ensuring that the YOLO model had not previously encountered them. This step was essential to avoid artificially improving detection accuracy and thus distorting the evaluation. The roadwork detection and localization system was tested on eleven construction sites\anon{ in Berlin}. For this evaluation, previously recorded rosbag files containing comprehensive sensor data from real-world drives were used to simulate the route, enabling systematic performance assessment. A major challenge, however, was the lack of reliable reference data for roadwork dimensions. Although this information should theoretically be available from Geoportal \anon{Berlin \cite{gdi}}, these records proved neither sufficiently precise nor complete to provide usable measurements for any of the investigated sites. To address this, an alternative method was adopted to establish the ground truth. The Velodyne Alpha Prime 3D point clouds (see Section \ref{sec:tech_infrastructure}) enable precise identification of individual roadwork objects based on their shapes. Relevant points were manually selected to determine the start, end, and most distant corner points of each roadwork (see Fig. \ref{Tiefenberechnung}). Along each identified edge, ten points were selected at the same horizontal position but at different heights. The mean of these vertically aligned samples was then used as the reference for that edge, thereby avoiding reliance on a single measurement. This approach reduces the influence of local irregularities in the point cloud. The resulting standard deviations ranged from 1 to 10 cm, depending on local point cloud density. At a distance of 15 m, the sensor’s intrinsic horizontal accuracy is about 5 cm (see Section \ref{sec:tech_infrastructure}). Ground truth measurement uncertainty therefore lies between 5 and 10 cm. These ground truths were then compared against the roadwork detection system outputs over three test runs. Based on nearly 100 test measurements, the mean Euclidean distance between the ground truth points and their corresponding detection system outputs was 0.32 m, with a standard deviation of 0.14 m. This demonstrates that the system achieves an accuracy on the order of a few decimeters. It is quite possible that the system’s actual accuracy is even better than reported. During the evaluation, it was noticeable that the LUX LiDAR contour was often wider than the point cloud (see Fig. \ref{Panels_too_small_in_ibeo}), or that the contour started slightly ahead of the point cloud (see Fig. \ref{ibeo_versetzt}). These discrepancies could be explained by the fact that the LUX sensors measure closer to the ground and might capture the support bases of roadwork objects, leading to a systematic offset compared to the point cloud. 
While difficult to confirm definitively, this possibility was investigated by transforming the measurements into global UTM coordinates and visualizing them in QGIS. Fig. \ref{GT_test} shows that the system’s measurements (red) sometimes extend slightly further into the street compared to the ground truth points (blue), likely because of the support bases. This reinforces the argument that system measurements might be even closer to the actual dimensions, potentially within the manufacturer’s stated margin of error (10 cm, see Section \ref{sec:tech_infrastructure}).
\subsection{Application for Geoportal \anon{Berlin}} \label{better_geoportal}
\begin{figure}[tb]
	\centering
	\includegraphics[width=0.24\textwidth]{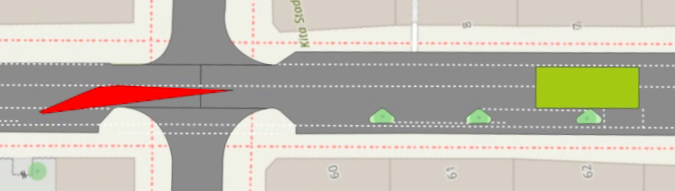}
	\caption{Comparison of roadwork measured by the system (red) and Geoportal \anon{Berlin \cite{gdi}} entry (green), registered half a block away from the actual position, therefore failing to show the intersection blockage}
	\label{roadwork_8_versetzt}
\end{figure}

All but two of the evaluated roadworks were officially listed in Geoportal \anon{Berlin \cite{gdi}}. However, none of these entries matched their true on-site locations as measured in this study. Either the geoportal failed to display that an entire road lane was blocked or the roadworks were registered roughly half a block away from their actual location (see Fig. \ref{roadwork_8_versetzt}). At times, the representation was overly broad, with the geoportal marking the entire motorway section as a construction site without highlighting the specific closed segments. Timing differences could explain some mismatches, but at least four of these roadworks were evaluated concurrently with the geoportal entries, confirming inaccuracies in the official data. Notably, in one example the roadwork was detected as early as November 2024, but the geoportal did not register any roadworks on that street until late April 2025, proving the information to be outdated. The study therefore demonstrates deficiencies in the data quality of Geoportal \anon{Berlin \cite{gdi}}. Automated detection systems like the one developed in this paper could significantly improve the accuracy and timeliness of roadwork mapping.
\subsection{System Performance} \label{system_performance}
Through various optimizations, including early loop termination, pre-calculation of transformations and skipping irrelevant objects, the execution speed of the roadwork detection process was significantly improved. Even in complex scenarios where over ten objects were processed in a single frame, the maximum computation time per cycle remained around 100 ms (10 Hz), with an average of 35 ms (28.5 Hz). The main bottleneck is managing the roadwork dictionary, as new detections must be correctly inserted among existing entries, which also need updating with the latest LiDAR data. Despite the dictionary sometimes exceeding fifty entries, processing stayed above 10 Hz. When no objects are detected by the CNN, processing time is dominated by YOLO inference, with the system achieving approximately 50 Hz. Overall, the system produces annotated images and roadwork measurements at approximately 11 fps. The output rate is constrained by the LiDAR update frequency, since the CNN processing is triggered only when new LiDAR data becomes available. Under high object load, the LiDAR updates at about 13 fps, making higher output rates impossible. Despite this limitation, the achieved 11 fps is sufficient for real-time roadwork detection.
\subsection{System Limitations and Future Work} \label{sec:limitations}
Several limitations became apparent during the evaluation of the roadwork detection system:

\textbf{Traffic cone detection:} While the CNN detects traffic cones effectively, the LUX LiDAR sensors struggle due to the cones' small size, limiting detection to when they are very close to the vehicle. This issue can only be resolved through the use of improved sensors.

\begin{figure}[tb]
	\centering
	\includegraphics[width=0.35\textwidth]{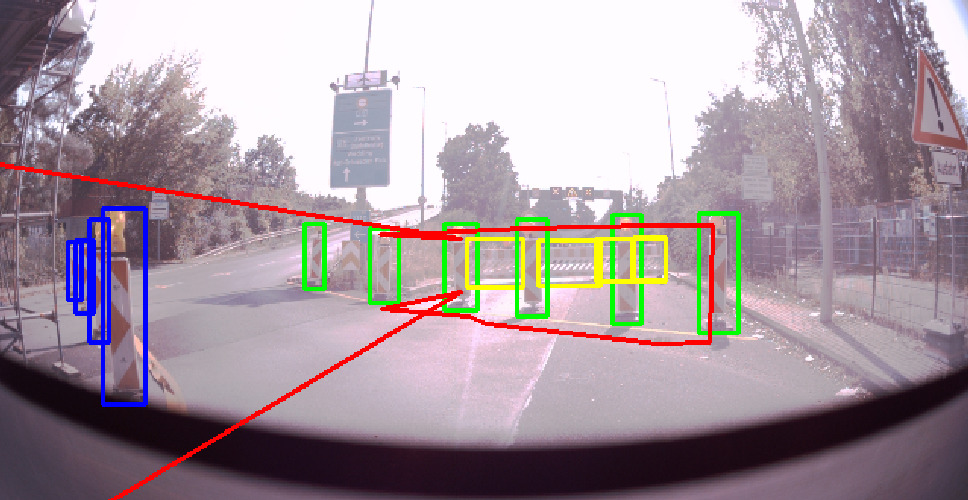}
	\caption{Vertical panel incorrectly associated with other panels across the road, causing a single roadwork to be formed due to the system's inability to account for road curvature}
	\label{kurven_fail}
\end{figure}

\textbf{Inability to handle curves:} The system sometimes incorrectly merges panels on opposite sides of the street in curves (see Fig. \ref{kurven_fail}). The solution could be to introduce a weighting algorithm utilizing the displayed direction of the vertical panels to group them correctly, thereby following the curvature of the road. Currently, once the vehicle’s direction aligns with the curve, this usually self-corrects, but can briefly confuse the system.

\textbf{Incomplete site measurement:} The system cannot yet capture the entire roadwork area, as objects at the end may be missed when other roadwork objects obstruct them (see Fig. \ref{Tiefenberechnung}). One potential improvement is to use the rear LUX LiDAR sensors combined with a backwards facing camera to measure previously hidden areas as the vehicle passes, helping to detect the full extent of the site.
\section{Conclusion}
The roadwork detection system developed in this paper can detect, localize and measure roadworks with an accuracy of at least 0.5 and likely closer to 0.1 meters. This was achieved by adapting a US roadwork dataset \cite{american_roadwork_dataset} for German traffic and creating a new dataset from test drives of the \anon{“MadeInGermany” prototype} \cite{prototype_vehicle}. Both datasets were used to train a YOLO CNN, whose predictions were fused with LiDAR measurements to detect and localize roadwork objects. These objects are aggregated in real-time as the vehicle passes by, forming coherent roadwork regions whose outlines are measured and converted into global UTM coordinates for logging and later use. It was further demonstrated that the road construction sites registered in Geoportal \anon{Berlin \cite{gdi}} are not sufficiently accurate, as some existing roadworks are missing and the recorded locations are consistently imprecise. The system presented in this paper could help to remedy this shortcoming and its deployment, along with the data collected during vehicle operation, could therefore be of great interest to road traffic authorities. If it is possible to implement all, or at least some, of the extensions proposed in Section \ref{sec:limitations}, it is conceivable that the system could continue to lay the foundation for maneuvering an autonomous vehicle around a road construction site. Individual roadwork objects are already detected and successfully aggregated into coherent roadworks by the system and their positions can be localized with sufficient accuracy by the LUX LiDAR sensors in order to allow the vehicle to navigate around them. With these fundamentals, it should be possible to determine the optimal driving trajectory, taking the construction site into account. As road construction sites pose an increased accident risk for motorists \cite{work_zone_crashs}, this system could not only reduce congestion and slow traffic, but also help prevent rear-end collisions or even save lives.
\anon{\section*{Acknowledgment}
The research leading to these results has received funding from the project KIS'M (FKZ: 45AVF3001F), funded by the German Federal Ministry for Digital and Transport (BMDV) program: ``A future-proof, sustainable mobility system through automated driving and networking''.}
\printbibliography

@book{cross_val_error,
  title={The Elements of Statistical Learning, Data Mining, Inference and Prediction, Second Edition},
  author={Hastie, T. and Tibshirani, R. and Friedman, J.},
  publisher={Springer New York},
  year={2009},
  isbn ={978-0-387-84858-7},
  pages = {pp. 241–257},
}

@online{rsa_allgemeines,
  author       = {StVZO},
  title        = {Richtlinien für die Sicherung von Arbeitsstellen (RSA) - Allgemeines},
  year         = {2022},
  url          = {https://www.stvzo.de/rsa/teil-a},
  urldate      = {2025-01-18},
}

@online{rsa_mainpage,
  author       = {U. Korsch},
  title        = {RSA-online - die Website zur RSA 21},
  year         = {2023},
  month	       = {3},
  url          = {http://www.rsa-online.com/index.htm},
  urldate      = {2025-07-21}
}

@unpublished{american_roadwork_dataset,
  author = {Ghosh, A. and Tamburo, R. and Zheng, S. and Alvarez-Padilla, J. R. and Zhu, H. and Cardei, M. and Dunn, N. and Mertz, C. and Narasimhan, S.},
  title = {ROADWork Dataset: Learning to Recognize, Observe, Analyze and Drive Through Work Zones},
  note = {Unpublished},
  year = {2024},
  month = {June},
  url = {https://www.cs.cmu.edu/~roadwork/}
}

@online{delays_US,
	title = {Making Work Zones Work Better},
	author = {FHWA},
	year = {2004},
	month = {4},
	url = {https://ops.fhwa.dot.gov/aboutus/one_pagers/wz.htm},
	urldate = {2025-04-18}
}

@online{work_zone_crashs,
  author = {National Safety Council},
  title = {Motor Vehicle Safety Issues - Work Zones},
  url = {https://injuryfacts.nsc.org/motor-vehicle/motor-vehicle-safety-issues/work-zones},
  urldate = {2025-04-18}
}

@article{BDD100K,
  author={Yu, F. and Chen, H. and Wang, X. and Xian, W. and Chen, Y. and Liu, F. and Madhavan, V. and Darrell, T.},
  journal={Proc. IEEE Conf. Comput. Vis. Pattern Recognit}, 
  title={BDD100K: A Diverse Driving Dataset for Heterogeneous Multitask Learning}, 
  year={2020},
  pages={2633-2642}
}

@article {mapillary,
	title = {The Mapillary Vistas Dataset for Semantic Understanding of Street Scenes},
	author = {Neuhold, G. and Ollmann, T. and Bulo, S. R. and Kontschieder, P.},
	journal = {Proc. IEEE Int. Conf. Comput. Vis.},
	year = {2017},
	issn = {2380-7504},
	pages = {5000-5009},
	month = Oct
}

@online{prototype_vehicle,
  author       = {AutoNOMOS Labs},
  title        = {MadeInGermany},
  url          = {https://autonomos.inf.fu-berlin.de/vehicles/made-in-germany},
  urldate      = {2025-04-20},
}

@online{gdi,
  author = {Senatsverwaltung für Stadtentwicklung Bauen und Wohnen Berlin},
  title = {Planbare Ereignisse im öffentlichen Straßenland},
  url = {https://gdi.berlin.de/services/wfs/planb_ereignisse?REQUEST=GetCapabilities&SERVICE=wfs},
  urldate = {2025-05-05}
}

@article{work_zone_detection,
  author={Shi, W. and Rajkumar, R. R.},
  journal={Proc. IEEE Int. Conf. Intell. Transp. Syst.}, 
  title={Work Zone Detection For Autonomous Vehicles}, 
  year={2021},
  pages={1585-1591}
}

@article{Segmentation_of_Work-Zone_Scenes,
author = {V. Sundharam and A. Sarkar and A. Svetovidov and J. S. Hickman and A. L. Abbott},
title ={Characterization, Detection, and Segmentation of Work-Zone Scenes From Naturalistic Driving Data},
journal = {Transp. Res. Rec.},
volume = {2677},
number = {3},
pages = {490-504},
year = {2023}
}

@article{SHRP_2_NDS,
  author={Abodo, F. and Rittmuller, R. and Sumner, B. and Berthaume, A.},
  journal={Proc. Int. Conf. Mach. Learn. Appl.}, 
  title={Detecting Work Zones in SHRP 2 NDS Videos Using Deep Learning Based Computer Vision}, 
  year={2018},
  pages={679-686}
}

@article{Highway_Workzones,
  author={Seo, Y. and Lee, J. and Zhang, W. and Wettergreen, D.},
  journal={IEEE Trans. Intell. Transp. Syst.}, 
  title={Recognition of Highway Workzones for Reliable Autonomous Driving}, 
  year={2015},
  volume={16},
  number={2},
  pages={708-718}
}

@article{anticipate_roadworks,
  author={Mathibela, B. and Osborne, M. A. and Posner, I. and Newman, P.},
  journal={Proc. IEEE Intell. Transp. Syst. Conf.}, 
  title={Can priors be trusted? Learning to anticipate roadworks}, 
  year={2012},
  pages={927-932}
}

@article{Construction_Sites_on_Motorways,
  author={Kunz, P. and Schreier, M.},
  journal={Proc. IEEE Intell. Veh. Symp.}, 
  title={Automated Detection of Construction Sites on Motorways}, 
  year={2017},
  pages={1378-1385}
}

@online{yolo11_ultralytics,
  author = {G. Jocher and J. Qiu},
  title = {Ultralytics YOLO11},
  url = {https://github.com/ultralytics/ultralytics},
  year = {2024}
}

@article{Mask-RCNN,
  author={He, K. and Gkioxari, G. and Dollár, P. and Girshick, R.},
  journal={Proc. IEEE Int. Conf. Comput. Vis.}, 
  title={Mask R-CNN}, 
  year={2017},
  pages={2980-2988}
}
\end{document}